\documentclass[12pt]{article}
\usepackage{bqf-paper}
\usepackage{algorithm}
\usepackage{algpseudocode}
\usepackage{amsmath}
\usepackage[utf8]{inputenc} 
\usepackage[T1]{fontenc}    
\usepackage{hyperref}       
\usepackage{url}            
\usepackage{booktabs}       
\usepackage{amsfonts}       
\usepackage{nicefrac}       
\usepackage{microtype}      
\usepackage{xcolor}         
\usepackage{tikz}
\usetikzlibrary{arrows.meta,positioning,calc,decorations.pathreplacing,shapes.geometric}
\usepackage{caption}
\usepackage[table]{xcolor}
\usepackage{listings}

\newcommand{\ignore}[1]{{}}

\title{\paperTitleSmallCaps{
Long-Horizon AI Research for Grothendieck Constant: \\ A Case Study in Human--AI Mathematical Collaboration
}}

\author{
  \normalsize
  \setlength{\tabcolsep}{15pt}
  \begin{tabular}{ccc}
    Alan Li\thanks{Equal contribution.} & Rahul Saha\footnotemark[1] & Anton Xue \\
    \small\texttt{alanli@cs.utexas.edu} & \small\texttt{rahul.saha@utexas.edu} & \small\texttt{anton.xue@utexas.edu} \\[3ex]
    Swarat Chaudhuri & Adam Klivans & Pravesh K. Kothari \\
    \small \texttt{swarat@cs.utexas.edu} & \small\texttt{klivans@cs.utexas.edu} & \small\texttt{kothari@cs.princeton.edu} \\[3ex]
     & Raghu Meka &  \\
     & \small\texttt{raghum@cs.ucla.edu} & \\[3ex]
  \end{tabular}
}

\date{}
\begin{document}

\maketitle


\abstract{
AI agents are increasingly used in mathematics research, but it is often unclear how to use
them effectively. Long-horizon mathematical research in particular involves planning and execution over
a research state that evolves over an extended timescale, thereby introducing notable challenges related to limited context and scarce rewards.
Towards this, we present an extensive case study of how AI was used to improve bounds on the Grothendieck constant $K_G$, which captures the hardness between combinatorial problems and their continuous relaxations.
Specifically, while the precise value of $K_G$ is not known, we recently tightened the best known bounds to
\[
    \frac{6\pi}{11}
    \;\le\;
    K_G
    \;\le\;
    \frac{\pi}{2\log(1+\sqrt2)} - 10^{-4}.
\]
Crucially, these improvements were achieved using an AI research system spanning weeks that could arrive at insights deemed novel by domain experts.
We give a detailed discussion of our experience using AI for mathematics research, particularly touching upon its strengths and weaknesses, as well as our experience with creating ideal conditions for AI to arrive at breakthrough insights.
The mathematical results are presented and proved in a companion paper~\cite{saha2026new}.
}

\newpage

\tableofcontents


\section{Introduction}
\label{sec:intro}

The mathematical capabilities of AI systems have grown at a remarkable pace.
Within a few years, they have progressed from solving grade-school word problems to matching the best human performers on competition and advanced mathematics benchmarks~\cite{hubert2026alphaproof, imo-gold, glazer2024frontiermath}.
Mathematics research, however, is a \emph{long-horizon} task. This involves sustained reasoning over extended periods of time, testing out different approaches, and maintaining a large and evolving research state of progress, failures, and future directions. Moreover, the goals are often complex and require a lot of intermediary work whose significance might only be apparent much later, and the goals themselves might change based on new information discovered. It is still an open question how to use AI effectively at this horizon.


In this work, we provide a detailed case study of our experience using AI operating at the level of a research \emph{program}: deciding what to attempt next, retaining what failed and why, and converting an accumulation of failures into new mathematics.
We engineered an AI research system asynchronously steered by human operators.
We directed it at a concrete open problem: the Grothendieck constant $\KG$, which quantifies the hardness threshold between hard combinatorial optimization problems and their efficient continuous relaxations, and whose value has been open since 1953. We ran this continuously from June 16 to July 24, 2026, with over 240 sessions and 2000 reasoning model calls.

The collaboration tightened the best known bounds on both sides,
\[
    1.7135\ldots
    \;=\;
    \frac{6\pi}{11}
    \;\le\;
    \KG
    \;\le\;
    \frac{\pi}{2\log(1+\sqrt2)} - 3.47\times10^{-4}
    \;=\;
    1.7818\ldots,
\]
The upper bound comes from a new asymptotic framework for rounding algorithms, and the lower bound from the first argument that bounds $\KG$ from below without constructing a hard instance.
Central steps of this mathematics originated with the AI system and were judged novel by domain experts; all results stated as theorems have been independently verified by the authors, and complete proofs appear in the companion paper~\cite{saha2026new}.
Our contributions are as follows.
\begin{enumerate}[label=(\arabic*), leftmargin=*]
    \item \textbf{Novel approaches for a longstanding open problem.}
    We give an expository account of the Grothendieck constant and of the results in \cref{sec:grothendieck}.
    The upper bound introduces \emph{limiting Krivine schemes}, which enlarge the space of rounding algorithms studied since Krivine's work.
    The lower bound converts obstructions on all rounding schemes into lower bounds on $\KG$; it is the first lower bound on the constant that does not proceed by constructing a hard instance.

    \item \textbf{A methodology for long-horizon AI research.}
    We document the system and outcomes in \cref{sec:ai-methodology}: its division of labor between a reasoning model and a coding agent, its file-based memory, its internal verification protocol, and its channel for asynchronous human steering.

    \item \textbf{A case study with an analysis of capabilities.}
    We reconstruct how the lower bound was found in \cref{sec:ai-case-study} and analyze the full record in \cref{sec:ai-discussion}.
    The record shows a consistent asymmetry: the system was strong at technical execution, but substantially less reliable at research judgement and at maintaining an accurate research state.
\end{enumerate}

\paragraph{Related and recent work.}
Recent successes of AI in mathematics, including on open problems, vary
in method and in the role of the human. 
Evaluator-guided search has
produced record constructions on problems whose candidate solutions can
be scored
automatically~\cite{funsearch,novikov2025alphaevolvecodingagentscientific,charton2024patternboostconstructionsmathematicslittle}. 
Frontier reasoning models have
also recently found constructions that disprove open conjectures. An
internal OpenAI model disproved the Erd\H{o}s unit-distance
conjecture~\cite{openai2026planarpointsets}, and 
Claude Fable~5 found a counterexample to the Jacobian
conjecture~\cite{jacobian}. 
Our work similarly lacks an automatic evaluator, but differs in its
focus on an extended research program in which progress requires not
only developing mathematical arguments, but also deciding which
intermediate results and directions are worth pursuing.
Concurrent to this work, OpenAI has also announced solutions to ten
open problems~\cite{openai2026ten}. The core results were found autonomously by OpenAI's unreleased internal model, Astra. In contrast, our work explores how human mathematicians can continuously collaborate with publicly available models over the course of weeks on an extended research program.

\section{The Grothendieck Constant}
\label{sec:grothendieck}

We first describe the mathematical problem at a high level. The exposition here is deliberately informal; precise definitions and full proofs appear in the companion paper~\cite{saha2026new}.

\paragraph{A hard optimization problem and its relaxation.}
Consider the following discrete optimization problem: given a matrix $A=(a_{ij})\in\R^{m\times n}$, find sign vectors maximizing the bilinear form
\[
    \OPT(A)
    \;:=\;
    \max_{x\in\{\pm1\}^m,\; y\in\{\pm1\}^n}
    \;\sum_{i,j} a_{ij}\, x_i y_j.
\]
Intuitively, we are looking for the best pair of $\pm1$ labelings of the rows and columns of $A$; this quantity is closely related to the \emph{cut norm} of the matrix, a basic primitive in combinatorial optimization, and the problem is NP-hard in general.
A standard way to make it tractable is to relax each sign to a unit vector and each product to an inner product:
\[
    \SDP(A)
    \;:=\;
    \max_{u_i, v_j \in S^{d-1}}
    \;\sum_{i,j} a_{ij}\, \ip{u_i}{v_j},
\]
where the maximum now runs over unit vectors in any dimension $d$.
This relaxation is a semidefinite program (SDP) and can be solved in polynomial time.
Since every sign is a one-dimensional unit vector, $\OPT(A)\le\SDP(A)$: the relaxation can only overshoot.
The basic question is by how much.

\begin{figure}[t]
\centering
\includegraphics[width=0.9\textwidth]{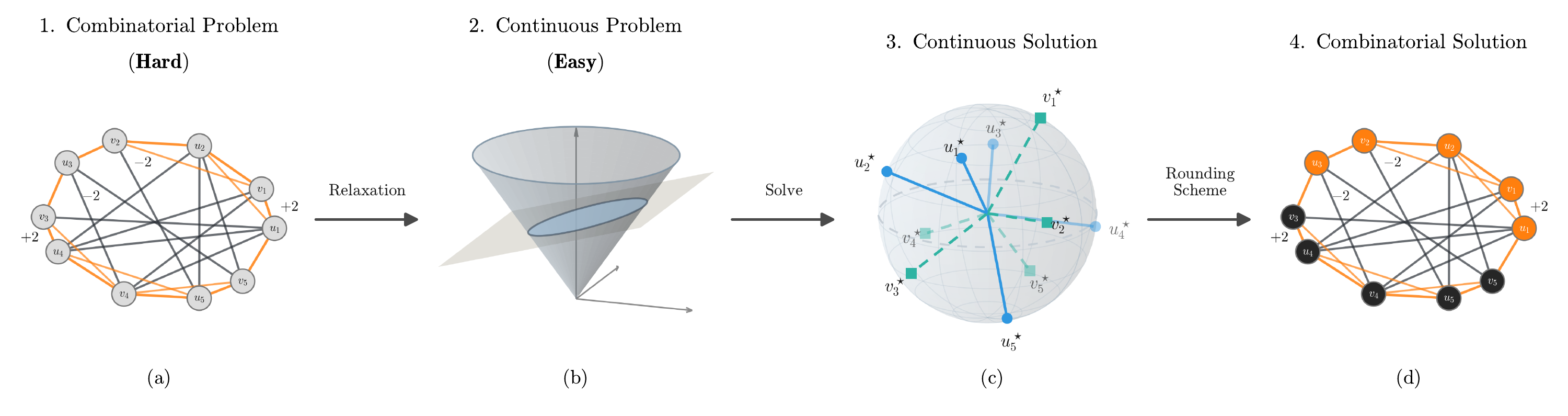}
\caption{\textbf{$K_G$ quantifies how well continuous methods can solve discrete problems.}
For illustrative purposes, we depict the standard relaxation-and-rounding pipeline: starting from a discrete combinatorial optimization problem \emph{(a)}, we pass to its efficiently solvable SDP relaxation \emph{(b)}, obtain an optimal vector solution \emph{(c)}, and round those vectors back to signs \emph{(d)}.
The Grothendieck constant $\KG$ is the worst-case ratio between the relaxed and discrete optima.
}
\label{fig:groth-pipeline}
\end{figure}

Grothendieck's inequality~\cite{Gro53,LP68} answers this question: the overshoot is bounded by a universal constant.
That is, there exists $K<\infty$, independent of $A$, $m$, $n$, and the dimension $d$, such that
\[
    \OPT(A)
    \;\le\;
    \SDP(A)
    \;\le\;
    K\cdot\OPT(A)
    \qquad
    \text{for every matrix } A.
\]
The \emph{Grothendieck constant} $\KG$ is the smallest such $K$.
In modern language, $\KG$ is exactly the worst-case integrality gap of the canonical SDP relaxation of the bilinear optimization problem above.
\cref{fig:groth-pipeline} illustrates the two optimization problems and the rounding step that connects them.

Grothendieck's inequality is a foundational result well beyond this algorithmic reading: it originated in functional analysis, where it is central to the geometry of Banach spaces and harmonic analysis~\cite{pisier2011grothendieckstheorempastpresent,KN11}, it underlies constant-factor approximation algorithms for cut norms~\cite{AN06}, and it governs the maximal advantage of quantum over classical correlations in Bell-type experiments~\cite{Tsi85,CHTW04}.
Despite this breadth of connections, and more than seventy years of study, the exact value of $\KG$ remains unknown.

\begin{figure}[t]
\centering
\includegraphics[width=0.85\textwidth]{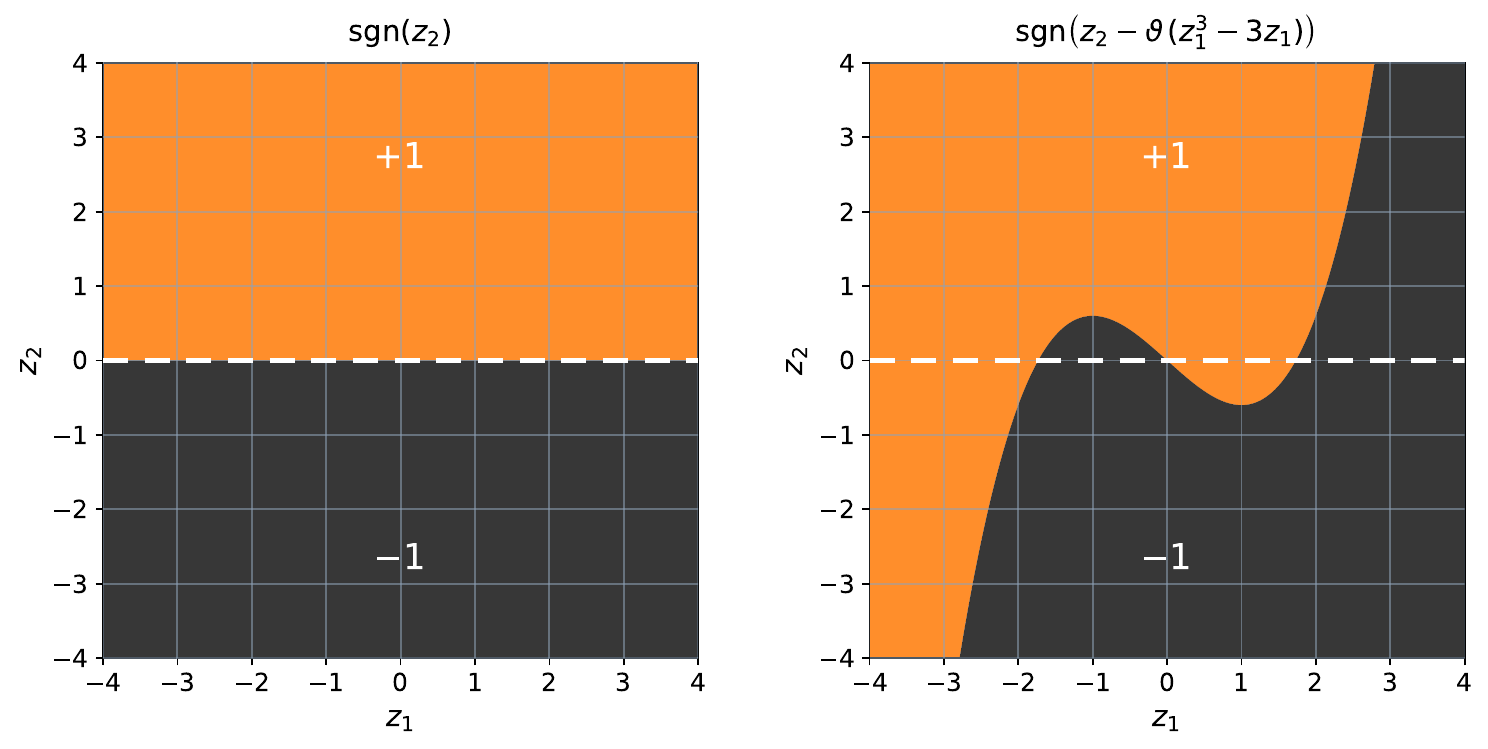}
\caption{\textbf{Krivine schemes are used to discretize continuous solutions.}
The main idea is to partition the ambient space into two regions, labeled $+1$ and $-1$, and assign each projected SDP vector according to the region in which it lands.
\emph{Left:} the half-space partition underlying hyperplane rounding and Krivine's classical bound.
\emph{Right:} a partition with a curved boundary, of the kind used in the companion paper.}
\label{fig:partitions}
\end{figure}

\paragraph{Rounding schemes as partitions.}
Every upper bound on $\KG$ is, at heart, an algorithm: a \emph{rounding scheme} that converts the unit vectors of an SDP solution back into signs while provably retaining a $1/K$ fraction of the objective.
The schemes relevant to this paper have a simple geometric description.
A \emph{Krivine scheme} is a pair of partitions of $\R^k$ into a $+1$ region and a $-1$ region, encoded by odd sign functions $f,g:\R^k\to\{\pm1\}$ (see \cref{fig:partitions}).
To round, the algorithm maps each SDP vector to a random Gaussian point in $\R^k$, arranged so that the points of $u_i$ and $v_j$ are correlated according to the inner product $\ip{u_i}{v_j}$, and then labels each vector by the region its point lands in: $x_i = f(X_i)$ and $y_j = g(Y_j)$.
The simplest instance takes both partitions to be the same half-space, $f(z)=g(z)=\sgn(z_2)$.
This is called the random hyperplane rounding: label each vector by the side of a random hyperplane it falls on.

\paragraph{The normalized correlation function.}
The quality of a Krivine scheme is governed by a single analytic object.
Let $X,Y$ be standard Gaussian points in $\R^k$ whose coordinates are pairwise correlated: $\E[X_iY_i]=t$ for each $i$.
The \emph{normalized correlation function} of the scheme is
\[
    H(t)
    \;:=\;
    \frac{\pi}{2}\,\E\bigl[f(X)\,g(Y)\bigr],
\]
which measures how correlated the two output labels are when the underlying inputs have correlation $t$.
This function is the central object in the analysis of Krivine schemes: the entire performance of a scheme can be read off from it, and, speaking informally, the closer $H$ is to a straight line with a large slope, the better the scheme rounds.

\vspace{1em}
For the half-space partition, a classical identity of Grothendieck gives $H(t)=\arcsin t$, and Krivine's analysis of this arcsine nonlinearity yields his celebrated bound~\cite{Kri77}
\[
    \KG
    \;\le\;
    \frac{\pi}{2\log(1+\sqrt2)}
    \;=\;
    1.7822\ldots
\]
Krivine conjectured that this value is optimal.
The conjecture stood for over three decades, until Braverman, Makarychev, Makarychev, and Naor disproved it in 2011 by mixing hyperplane rounding with a carefully chosen two-dimensional scheme; their argument establishes the strict inequality $\KG<\pi/(2\log(1+\sqrt2))$ but does not quantify the gap~\cite{braverman2011grothendieckconstantstrictlysmaller}.
Naor and Regev later proved a striking converse: such mixed Krivine schemes are asymptotically \emph{optimal}, meaning that as the dimension of the scheme grows, they achieve approximation ratios arbitrarily close to the true value of $\KG$~\cite{naor2014krivine}.
The search for $\KG$ can therefore, in principle, be carried out entirely within this one family of partitions.
Building on this, recent works of Heilman~\cite{heilman2026upperboundgrothendiecksconstant} and of a subset of the authors~\cite{LiSahaXueChaudhuriKlivansKothariMeka2026} independently obtained the first explicit numerical improvements over Krivine's bound, on the order of $10^{-5}$.

\paragraph{Limiting Krivine schemes.}
The companion paper enlarges the space of previously known schemes.
A \emph{limiting Krivine scheme} is obtained as a limit of classical Krivine schemes of growing dimension; it is again described by a pair of partitions, and it inherits the rounding guarantees of the schemes converging to it.
This enlarged search space is where the upper bound result lives.

\paragraph{Lower bounds are hard instances.}
Lower bounds on $\KG$ have historically come from the complementary direction: exhibiting an explicit matrix $A$ for which $\SDP(A)$ provably exceeds $\OPT(A)$ by a large factor.
Grothendieck's own work implies $\KG\ge\pi/2$, and a high-dimensional Gaussian construction found independently by Davie and Reeds gives $\KG\ge1.6769\ldots$~\cite{Davie1984LowerBoundKG,Reeds1991LowerBoundKG}.
That bound stood for more than four decades, until two recent works improved it by $10^{-26}$ and $10^{-12}$, respectively~\cite{heilman2026lowerboundgrothendiecksconstant,jones2026grothendieckconstantstrictlylarger}.
The state of the art entering this work was thus
\[
    1.6769\ldots \;\le\; \KG \;\le\; 1.7822\ldots,
\]
an interval wide enough that even the tenths digit of $\KG$ was unknown.

\paragraph{The Results of the Companion Paper.}
\label{sec:companion-results}
The companion paper~\cite{saha2026new} proves two results, one for lower bound and one for upper bound.
We state them here in abridged form; the rest of this paper needs only the statements and the shape of the arguments.

\begin{theorem}[Upper bound, abridged]
\label{thm:ub-abridged}
\label{thm:cubic-quintic-upper-bound}
There is an explicit limiting Krivine scheme, the \emph{cubic--quintic scheme}, which can be used to show
\[
    \KG
    \;\le\;
    \frac{\pi}{2\log(1+\sqrt2)} - 3.47\times10^{-4}.
\]
\end{theorem}

The scheme's partitions have cubic boundaries of the kind shown in \cref{fig:partitions} (right).
All previous constructions, including the two recent $10^{-5}$-scale improvements, were fixed low-dimensional schemes; this is the first improvement obtained by letting the dimension grow, and it answers affirmatively a question of Braverman et al.~\cite{braverman2011grothendieckconstantstrictlysmaller} on whether higher dimension helps.

\begin{theorem}[Lower bound, abridged]
\label{thm:lb-abridged}
The correlation function $H(t)=b_1t+b_3t^3+\cdots$ of every Krivine scheme satisfies the constraint
\begin{equation}
\label{eq:affine-coefficient-inequality}
    b_3 \;\ge\; 2b_1-\frac{11}{6}.
\end{equation}
Combined with the optimality theorem of Naor and Regev~\cite{naor2014krivine}, this implies
\[
    \KG \;\ge\; \frac{6\pi}{11} \;=\; 1.7135\ldots
\]
\end{theorem}

The interesting feature of \cref{thm:lb-abridged} is the reversal of strategy.
Instead of constructing a hard instance, the proof establishes a ceiling on the performance of \emph{every} scheme. 
This is the first lower bound on $\KG$ that does not proceed by constructing a gap instance.
The proof reduces the constraint on high-dimensional partitions to a small set of one-dimensional Gaussian inequalities.

Together, the two theorems give
\[
    \frac{6\pi}{11}
    \;\le\;
    \KG
    \;\le\;
    \frac{\pi}{2\log(1+\sqrt2)} - 3.47\times10^{-4},
\]
determining the tenths digit of $\KG$ to be $7$.
The remainder of this paper concerns how these results were found: the AI research system and its methodology, the run itself, and what the record shows about current AI systems as mathematical collaborators.

\hypersetup{
  colorlinks=true,
  linkcolor=blue!55!black,
  urlcolor=blue!55!black,
  citecolor=blue!55!black
}

\setlength{\parindent}{0pt}
\setlength{\parskip}{6pt}
\setlist[itemize]{leftmargin=1.6em,itemsep=2pt,topsep=3pt}
\setlist[enumerate]{leftmargin=1.6em,itemsep=2pt,topsep=3pt}

\section{AI Methodology and Analysis}
\label{sec:ai-methodology}

The lower bound results in this paper were obtained in collaboration with an AI research system. The upper bound
construction  (Theorem~\ref{thm:cubic-quintic-upper-bound})
predate the current system and were found through conversation with GPT-5.5-Pro. The lower bound $\KG\ge6\pi/11$ was
discovered and first proved by the system, and we have since verified the
argument ourselves, and the proof in the companion paper~\cite{saha2026new} is our revision of the
system's write-up. 

The system also produced further improvements on both
sides of the problem, which we report in Section~\ref{sec:ai-results} but do
not state as theorems, because their certificates have not yet been human-verified. 



\paragraph{AI models and compute.}
The run used two frontier language models: a reasoning model (OpenAI
GPT-5.5-Pro at maximal reasoning effort, replaced mid-run by GPT-5.6-Sol)
and a coding agent (Anthropic Claude Code, running Claude Opus and later
Claude Fable~5). Floating-point searches
ran on a single four-GPU node, and every Arb-certified computation was redone in interval arithmetic on CPU.

\paragraph{The AI research system.}
The design of a research system depends on the problem being explored. We used a \emph{harness} to build the system: a small collection of AI models, with software tools they can use, attacking a single research goal. The first design decision is which models and tools to include, and it is problem-specific. Our problem depends heavily on reliable numerics: both bounds are certified by rigorous interval arithmetic, and many hypotheses can be strength-tested or rejected via numerical experiments. A different problem might instead require a symbolic or numerical package such as Sage; we use the software library Arb~\cite{johansson2017arb} for our problem. 

Our harness has four components.
\begin{enumerate}
    \item A \textit{reasoning model} does the mathematics in natural language: it chooses what to work on, develops arguments, proposes experiments, and tests the resulting claims. We used a general-purpose reasoning model (GPT-5.5/5.6). In our experience, its mathematical strength was the single largest factor in how well the harness performed.
    \item A \textit{coding agent} does the computation: it writes, runs, and verifies the programs the reasoning model asks for. We used Claude Code (Opus 4.7/Fable) for the coding agent. For the upper bound it evaluated candidate rounding schemes, and for the lower bound it certified the one-dimensional inequality at the heart of the proof.
    \item A \textit{bulletin} is a file that carries instructions from the human operators to the harness.  The agents periodically read it as they work; this is how a human steers the research without halting it.
    \item A \textit{session report} is the harness's long-term memory.  The report contains code, proofs, and prose explaining the mathematics, that later work reads in order to continue. This allows progress to accumulate without clogging the context of every reasoning model call.
\end{enumerate}

\paragraph{Run configuration.}
The harness runs as a loop of \emph{sessions}. 

In each session, the harness starts by calling the reasoning model with the session reports and bulletin as context, and asks it to come up with a research direction. Then, the harness calls the reasoning model a few times afterwards with the research direction, and asks it to think theoretically about how it plans to execute the research direction. This step was added to ensure the reasoning model did not have myopic view of the research program, which we had empirically observed before. Finally, for the remaining reasoning model calls in the session (up to 12 maximum), it was allowed to ask to coding agent to execute computations and run experiments. Up to five sessions run in parallel so that different proof strategies can be pursued simultaneously. We initialized the first session with relevant literature and our work on the cubic-quintic scheme.

The human operators periodically reviewed the harness outputs by reading and summarizing the session reports. The appropriate frequency of these check-ins depends on the problem. 

\paragraph{Results.}
\label{sec:ai-results}
The run spanned June~16 to July~24, 2026. It comprised
roughly 240 research sessions. The reasoning model was called 2{,}091
times, consuming about 152 million tokens at an estimated \$5{,}400 in
API cost; the coding agent ran on a flat-rate subscription (Claude Max).
Table~\ref{tab:run} collects these figures, and
Table~\ref{tab:timeline} gives the timeline of the bounds.

\begin{table}[t]
\centering
\small
\begin{tabular}{@{}ll@{}}
\toprule
Duration & June 16 -- July 24, 2026 (main phase through July 4)\\
Research sessions & $\approx240$\\
Reasoning model & GPT-5.5-Pro, later GPT-5.6-Sol; maximal effort\\
Usage and cost & 2{,}091 calls; $\approx152$M tokens; $\approx\$5{,}400$\\
Coding agent & Claude Code (Opus, later Fable 5); subscription\\
Human steering  $\approx40$ dated directives\\
\bottomrule
\end{tabular}
\caption{\textbf{The run at a glance.} Aggregates computed from the harness
telemetry. Dollar figures are list-price estimates; coding-agent usage was
only partially metered.
}
\label{tab:run}
\end{table}

\begin{table}[t]
\centering
\small
\begin{tabular}{@{}lp{0.76\textwidth}@{}}
\toprule
Sessions & Event\\
\midrule
1 & run begins from the interval $1.6769566742\le\KG\le1.7818666070$
(the upper end is the cubic--quintic bound of this paper)\\
18 & upper-bound search plateaus; the operators direct a pivot to the
lower bound\\
44 & numerical upper bound $1.7802243$ recorded (later withdrawn)\\
55--57 & affine reframe; $\KG\ge6\pi/11$ derived\\
58--64 & $6\pi/11$ proof system-tested; first \LaTeX{} write-up\\
92, 117 & lower bounds $27\pi/49\approx1.7311$ and
$51\pi/92\approx1.7415$ system-tested\\
follow-up & the session-44 value found unsupported by machine testing and
withdrawn\\
follow-up & upper bound $1.781801841033$ system-tested\\
follow-up & upper bound $1.7813319810625639$ system-tested;
adversarial review passed\\
\bottomrule
\end{tabular}
\caption{\textbf{Timeline of the bounds produced during the run.}
system-tested means accepted by the verification protocol of
Section~\ref{sec:ai-methodology}. The last three events fall in the short
follow-up phases, whose sessions we do not cite individually. Of the
run's results, only $\KG\ge6\pi/11$ has additionally been verified by the
authors; it is the result stated as a theorem in this paper.}
\label{tab:timeline}
\end{table}

The run's principal outcome, and the one this paper states as a theorem,
is the lower bound $\KG\ge6\pi/11$, whose discovery
Section~\ref{sec:ai-case-study} reconstructs. Beyond it, the run produced
system-tested upper bounds of $1.781801841033$ and then
$1.7813319810625639$, improving on the cubic--quintic value, together with
the two stronger lower bounds listed in Table~\ref{tab:timeline}. We
report these as system-tested claims rather than theorems: each passed
the full internal protocol, but we have not yet verified their
certificates ourselves.
We study a complementary regime: sustained collaboration between human
mathematicians and publicly available models, in which humans repeatedly
steer the research program as its state, priorities, and mathematical
directions evolve.

\section{Case study: From a Failed Upper Bound Search to a Lower Bound}
\label{sec:ai-case-study}

The lower bound is the run's main discovery. We reconstruct it here in
outline, highlighting the human interventions that shaped it and the
mathematical contributions of the AI system once it had the right framing.

The lower bound began with a plateauing search for a better upper bound.
The system explored several variants of the limiting Krivine schemes.
Although the constructions differed, they repeatedly encountered the same
tradeoff: suppressing the unwanted nonlinear terms in the inverse majorant
series also weakened the leading term. The run recorded this pattern, but
continued to search for further constructions rather than treating the
repeated failures as evidence of a general obstruction.

The key human intervention was to recognize that the repeated failures likely reflected a general analytic obstruction, and that proving such an obstruction for mixed and limiting schemes would itself be mathematically valuable: by the optimality theorem of Naor and Regev~\cite{naor2014krivine}, it would imply a lower bound on $K_G$. The operators therefore redirected the system away from searching for new limiting Krivine schemes and asked it instead to synthesize the accumulated failures, identify their common structure, and attempt to prove a universal obstruction.

After further program-level directives, the system found the central
mathematical reframe and developed a complete proof
(see \cite[Part~1]{saha2026new}). It expressed the obstruction as an affine
inequality between the two leading coefficients of a scheme's correlation
function. Because this
inequality is preserved under averaging and limits, it extends to mixed and
limiting schemes. The system then reduced the required high-dimensional
estimates to one-dimensional inequalities and produced the computer-assisted
certificate, yielding
\[
    \KG \geq \frac{6\pi}{11}.
\]
The authors subsequently revised the exposition and independently verified
the proof. Appendix~\ref{app:ai-case-study} gives the
detailed chronology.

This case study illustrates the different roles of the human operators and the AI system throughout the run, which we elaborate on in the discussion below. 


\section{Discussion}
\label{sec:ai-discussion}

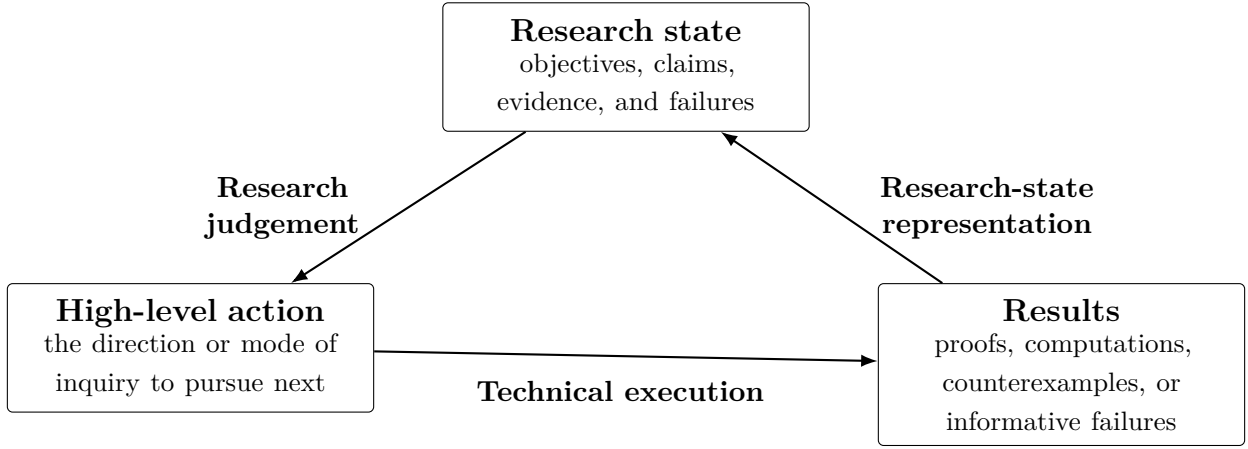
\begin{figure}[t]
  \centering
  \begin{tikzpicture}[
      box/.style={
        draw,
        rounded corners=2pt,
        align=center,
        minimum height=1.1cm,
        text width=0.245\textwidth,
        inner sep=6pt
      },
      flow/.style={-{Latex[length=2.2mm]}, thick},
      process/.style={font=\small\bfseries, align=center}
    ]
    \node[box] (state) {
      \textbf{Research state}\\[-1mm]
      \footnotesize objectives, claims, evidence, and failures
    };
    \node[box, below left=2.0cm and 0.9cm of state] (action) {
      \textbf{High-level action}\\[-1mm]
      \footnotesize the direction or mode of inquiry to pursue next
    };
    \node[box, below right=2.0cm and 0.9cm of state] (results) {
      \textbf{Results}\\[-1mm]
      \footnotesize proofs, computations, counterexamples, or informative failures
    };

    \draw[flow] (state) -- node[process, midway, left=14pt] {
      Research\\judgement
    } (action);
    \draw[flow] (action) -- node[process, midway, below=5pt] {
      Technical execution
    } (results);
    \draw[flow] (results) -- node[process, midway, right=14pt] {
      Research-state\\representation
    } (state);
  \end{tikzpicture}
  \caption{A simplified research cycle modelling the process of long-horizon
  mathematical research. The decomposition is intended as a systems-level
  description.}
  \label{fig:research-cycle}
\end{figure}

\paragraph{Structure of research process.}
Motivated by mathematical problem-solving frameworks such as~\cite{carlson2005problemsolving, schoenfeld1983episodes}, we present the following decomposition of long-horizon mathematical
research in the harness (see Figure~\ref{fig:research-cycle}). At any point in time, a research program has an active \emph{research state}. This is a compact representation of the complete history of the run for the purpose of choosing the next step, including current objectives, established and conjectural claims, evidence, failed approaches, open proof obligations, and unresolved choices. From this state, the
researcher uses \emph{research judgement} to select a high-level action. Such
an action might be to prove an auxiliary lemma, test a new construction,
compare several failures, search for a counterexample, reformulate the target,
or test an earlier claim. \emph{Technical execution} takes the high-level action, and then plans and executes the steps needed to produce results. Finally, those results must be distilled and incorporated into
an updated active represented state, in a way that is efficient and preserves all important information. The cycle then repeats.

The components of this research cycle have familiar precedents. Newell and Simon model problem
solving as the application of operators to states in a problem
space~\cite{newell1972human}; P\'olya separates understanding and planning
from execution and looking back~\cite{polya1957solve}; Schoenfeld emphasizes
\emph{control}, the executive decisions that select and monitor mathematical
strategies~\cite{schoenfeld1983episodes,schoenfeld1985mathematical}; and
Sch\"on describes professional inquiry as a reflective conversation in which
action changes the situation and may require it to be reframed
~\cite{schon1983reflective}. Figure~\ref{fig:research-cycle} is a minimal synthesis of these ideas, adapted to the architecture of the harness.

\paragraph{Performance across research functionalities.}
The system consistently exhibited an asymmetry in its capabilities between its local and global functionalities. It was exceptionally strong at
technical execution, and substantially less reliable at research judgement and research-state representation, the two functionalities that operate over the research program as a whole.

\noindent \textit{(1) Technical execution: strong.}
Once a high-level action was fixed, the system was
usually effective at thinking creatively and devising the mathematics necessary to advancing the action. It drew upon its vast range of
mathematical knowledge and problem-solving skills to develop lemmas and proofs, and it also effectively ran and interpreted experiments. A prominent example of this was how the system found and developed the current lower bound, as detailed in the case study. After the upper-bound failures were reframed as evidence for a
universal obstruction, the system developed the central proof of the lower
bound with comparatively little further intervention. This required a lengthy argument with many intermediate results, involving creative application of areas such as Gaussian harmonic analysis, geometric and functional inequalities, variational arguments, and finite-dimensional optimization.

\noindent \textit{(2) Research judgement: limited autonomy.}
Long-horizon research requires repeated decisions about what the program
should do next~\cite{tao2007good,thurston1994proof}. These decisions combine 
expected significance, generality, downstream use of
a result, information
gained from failure, and verification cost. The system did not reliably make these decisions at the program level. The initial
upper-bound search in the case study is a clear example: several distinct approaches had
failed for the same structural reason, yet the run opened another variation
rather than asking whether the common obstruction was universal, and could be proven as a theorem. Other failings of the system included not recognizing the value of a theoretical result over numerical evidence, and prioritizing a sharper restricted result that is not inherently interesting over a general result. The relevant limitation was not an inability to perform the required
reasoning. Once prompted to step back, synthesize, or reframe, the system
often did so effectively. It was the failure to recognize, from the evolving
state of the project, when such a change in reasoning mode was called for.
This resembles the distinction made by Mason and Spence~\cite{mason1999beyond} between possessing a strategy and
\emph{knowing to} activate it in the appropriate situation. 

\noindent \textit{(3) Research-state representation: fragile.}
This functionality became more difficult as the run grew. The project produced
far more intermediate mathematics than could be placed in a model's context
at once, so the harness maintained an archive of transcripts and experiments
together with a much smaller curated summary, rewritten at every merge, that
each new session read as its working memory. The withdrawn upper bound in session 44 (see Table~\ref{tab:timeline}) is a
clear case of failure. A fast numerical evaluator built in session~4 carried
an explicit caveat that its score was safe only for exploration; the caveat
did not survive later representations of the research state, and by
session~44 the uncertified score had set the record.
A feasibility criterion that would have invalidated the record was lost even earlier: the run proved it in session~8, but it disappeared in a later handoff and was eventually reproved from scratch by the test that withdrew the record 25 days later.
In both failures, the archive retained the original facts; it was the compressed state governing decisions that
failed.

\paragraph{Why is the AI good and bad at what it does?}
\label{sec:ai-origins}
We hypothesize the following explanation for the AI system's performance in each of these functionalities. There is a plausible asymmetry in both the training data and feedback received by the base model, where it has a lot more experience with local, technical execution over global, long-horizon reasoning and planning. Mathematical papers, textbooks, and solved problems provide many
examples of technical execution where definitions are used and results are proven. Moreover, AI models can be trained effectively on problems where the final answer or proof can be scored automatically~\cite{shao2024deepseekmath,hubert2026alphaproof}, which roughly corresponds to the skills required for strong technical execution we see in the run.

The training data and feedback available for research judgement is much rarer. Finished
mathematical papers rarely record the full process of discovery, including the abandoned approaches, the experiments performed, and decisions to pivot directions. Instead, they represent a polished final state, written for expositional clarity, rigor, and brevity. Lakatos's contrast between deductive exposition and the process of proofs,
criticism, and revision makes this explicit~\cite{lakatos1976proofs}; Thurston likewise argues that mathematical progress
and understanding are not captured by formal proofs alone~\cite{thurston1994proof}. Moreover, it is computationally much more difficult to run long-horizon mathematical research projects in the same training environments as the more elementary mathematical problems AI models are trained on. The reward design for this would also be more difficult, as reward signals are very sparse and success can be open-ended.

The weakness in research-state representation fits the same asymmetry.
Curation is a global functionality: deciding what the program will later need to
know is a prediction about its future, of the same kind as the judgements
above. The record that would teach the model this prediction (what was kept, what was dropped,
and which omission later mattered) is what finished exposition
omits. The hypothesis also predicts the direction of the failure we
observed. The active state was rewritten roughly two hundred times over the
run, and each rewrite draws on a training genre, the finished mathematical paper, that
states results and omits caveats, scope, and doubt. Therefore, iterated rewriting regresses toward that genre, and headline values survive retelling
more reliably than their evidential status, a bias documented in both human
and language-model
retelling~\cite{bartlett1932remembering,acerbi2023transmission,mohamed2025broken}.
The corpus records mathematics as product rather than process, and both
global functionalities need the process view.


\paragraph{Why we still need human collaboration.}
The decomposition is also instructive in classifying the human contribution in the run, which entered in three areas. First, the operators supplied the initial
research state, including the cubic--quintic framework and its certification
machinery, and a problem statement with candidate directions for both
bounds (Section~\ref{sec:ai-methodology}). Second, through about forty directives, they supplied
program-level research judgement: selecting and redirecting the
high-level action (the lower bound pivot of Section~\ref{sec:ai-case-study} is the
crucial instance) and fixing priorities such as generality over
numerical sharpness. Third, they
intervened in research-state representation, continuously changing the structure of the research-state representation, and changing the guidelines for what information to store and how. The technical execution was left almost entirely
to the system.

This contribution is itself mathematically significant. Recognizing when to switch to another approach, or when a common obstruction might be able to be generalized to a useful theorem are instances of the higher-level, largely intuitive
judgement that mathematicians acquire with
experience~\cite{tao2007good,thurston1994proof}.
The same holds for research-state representation: deciding which information may matter to future directions requires foresight and experience with mathematical research.

\paragraph{Future directions.}
Our analysis suggests two complementary directions. For AI research, improving research judgement and research-state representation may require training and evaluation on records of mathematics as a process, including failed approaches, strategic decisions, and evolving assessments of evidence. For mathematical practice, while these limitations persist, human expertise is likely to remain most valuable in these global functionalities: deciding when to persist or reframe a direction, and maintaining an accurate representation of accumulated progress.

\paragraph{Acknowledgements.} We thank Sabrina Reguyal for reviewing and providing helpful suggestions for the manuscript.

\bibliographystyle{amsalpha}
\bibliography{references}

\newpage
\appendix

\section{Extended Case Study: The Lower Bound}
\label{app:ai-case-study}


This is a more detailed chronology of the lower bound case study.


\paragraph{The plateau.}
The run began on the upper bound, searching the limiting-scheme space of our
preprint for schemes whose correlation function admits a large
inverse-majorant parameter $\gamma$, since an admissible $\gamma$ gives
$\KG\le\pi/(2\gamma)$. From the hyperplane value
$\gamma=\log(1+\sqrt2)\approx0.8814$ it reached $\gamma\approx0.882$, and
there it stalled: in sessions 12 through 17 it opened six 
mechanically distinct escape routes, and each failed against the same obstruction,
namely that cubic coefficient mass cannot be cancelled without also reducing
the linear coefficient. Each failure was recorded and tagged, and the run's
response was to begin a seventh variation of the same search.

\paragraph{First intervention: turn the failures into a theorem.}
The operators redirected the run in session 18. Instead of attempting another
escape, it was to synthesize the accumulated failures into a proof that no
scheme in the class does much better: a ceiling $\gamma\le\Gamma$ valid for
the full class, with mixtures and arbitrary dimension, converts through the
optimality theorem of Naor and Regev \cite{naor2014krivine} into the lower
bound $\KG\ge\pi/(2\Gamma)$. The intervention supplied no new mathematics:
the problem statement had suggested this dualization from the outset, and
the run had accumulated exactly the required evidence without making the
connection. Session 18 identified the Naor--Regev theorem as the correct
transfer, proved a first ingredient valid in every dimension, and located
the difficulty precisely: the ceilings it could prove covered too narrow a
class of schemes.

\paragraph{Progress in two dimensions.}
Within days the run had proved a cubic ceiling of strength
$\Gamma\approx0.8866$ for two-dimensional schemes, with the inequality certified in interval arithmetic, and had sharpened the
constant numerically to $\Gamma\approx0.8846$. The extension to arbitrary dimension and to mixtures, which the
duality requires, did not follow: fourteen sessions in the range 32--51
attacked it, each reducing the gap to a sub-lemma, proving the sub-lemma in
a restricted setting, and leaving the global assembly open, while the
restricted results accumulated in memory as apparent progress. Two useful
facts emerged alongside this loop: the run verified the normalization of the
Naor--Regev transfer against the original paper, and it computed that
improving the Davie--Reeds~\cite{Davie1984LowerBoundKG,Reeds1991LowerBoundKG} bound $1.6769566742$ requires a universal ceiling
only at $\Gamma\le0.93669$, far above the two-dimensional value, so a
general ceiling could be far from sharp and still give a new lower bound.

\paragraph{Second intervention: the general bound, through theory.}
The operators ended this loop with a second redirection, on two grounds:
refinements of the two-dimensional result were not the object of interest,
since only a ceiling for the full class dualizes; and the gap was
conceptual, so further implementation would not close it. The run was to
spend several sessions on theory alone. Session 55 mapped the dependency
structure of the preceding weeks and identified the recurring error,
restricted closures recorded as global progress. Session 56 then produced
the 
decisive reframe. Every failed transport had been nonlinear in
quantities that do not average when schemes are mixed; the Hermite
coefficients $(b_1,b_3)$ of the correlation function, by contrast, are
linear functionals of the scheme, so a constraint that is affine in
$(b_1,b_3)$ passes to mixtures and to coefficientwise limits with no further
argument. Choosing the constraint for transportability, and requiring
equality at the extremal hyperplane point $(b_1,b_3)=(1,\tfrac16)$, gives
the one-parameter family of affine constraints
\[
        b_3\;\ge\;(1+\lambda)\,b_1-\Bigl(\lambda+\tfrac56\Bigr),
        \qquad \lambda\ge\tfrac12 ,
\]
where $\lambda$ indexes the slope of the line through the hyperplane point.
Each member forces a barrier $\Gamma_\lambda=(\lambda+\frac56)/(1+\lambda)$
on the admissible parameter, hence a bound $\KG\ge\pi/(2\Gamma_\lambda)$
conditional on proving that member, and every $\Gamma_\lambda$ below
$0.93669$ improves on Davie--Reeds.

\paragraph{Third intervention: one certified rung before a better one.}
The tangent member $\lambda=\tfrac12$, worth $\KG\ge9\pi/16\approx1.767$,
requires a sharp chaos inequality that remains open, and once the family
existed the run's inclination was to work toward its strongest provable
member. The operators set the priority differently: what mattered was to
establish that the method yields any new lower bound at all, so the run was
to select a rung whose proof closes, certify it completely, and write it up,
with improvements inside the family treated as secondary. At $\lambda=1$ the
proof collapses to elementary structure: session 57 found that the
decomposition $h=(f+g)/2$, $k=(f-g)/2$ reduces the constraint, fiber by
one-dimensional fiber, to the ternary inequality,
whose closing case analysis is short, giving $\Gamma_1=\tfrac{11}{12}$ and
\[
        \KG\;\ge\;\frac{6\pi}{11}\;\approx\;1.7136 .
\]
Sessions 58--64 then subjected the chain to adversarial review, which
found and repaired the one substantive gap (the passage from finitely many
schemes to measurable mixtures, closed by a weak-$*$ lemma), a Jensen
inequality applied in the wrong direction, and a small defect in the
interval certificate's coverage, after which the run produced the first
complete write-up, about a day after the reframe. The stronger rungs
$27\pi/49$ and $51\pi/92$ of Table~\ref{tab:timeline} followed within a week
by the same method at smaller values of $\lambda$.

\end{document}